\documentclass{article}
\pdfoutput=1

    \PassOptionsToPackage{numbers, compress}{natbib}

\usepackage[preprint]{neurips_data_2024}

\usepackage[utf8]{inputenc} 
\usepackage[T1]{fontenc}    
\usepackage{hyperref}       
\usepackage{url}            
\usepackage{booktabs}       
\usepackage{amsfonts}       
\usepackage{nicefrac}       
\usepackage{microtype}      
\usepackage{xcolor}         

\usepackage{amsmath,bm}
\usepackage{enumitem}
\usepackage{pifont}
\usepackage{multirow}
\usepackage{makecell}
\usepackage{graphicx}
\usepackage{colortbl}
\usepackage{color}
\usepackage{wrapfig}
\usepackage{graphicx}
\usepackage{tabularx}
\usepackage{capt-of}
\usepackage{wrapfig}
\usepackage{subcaption}
\usepackage{graphicx}

\definecolor{lightblue}{cmyk}{0.1,0.0,0.02,0.02}

\title{Biomedical Visual Instruction Tuning \\ with Clinician Preference Alignment}

\author{
Hejie Cui$^{1,2}$\footnotemark[1], 
~~Lingjun Mao$^{3}$\thanks{These authors contributed equally to this work.}, 
~~Xin Liang$^{3}$, 
~~Jieyu Zhang$^{4}$, \\\bf
~~Hui Ren$^{5,6}$, 
~~Quanzheng Li$^{5,6}$, 
~~Xiang Li$^{5,6}$,
~~Carl Yang$^{2}$ \\
$^1$ Stanford University 
$^2$ Emory University 
$^3$ University of California, Berkeley \\
$^4$ University of Washington
$^5$ Massachusetts General Hospital
$^6$ Harvard Medical School
}

\usepackage{xspace}
\usepackage{natbib}
\newcommand{\ours}{{BioMed-VITAL}\xspace}
\newcommand{\oursv}{{BioMed-VITAL}}
\usepackage{pifont}
\usepackage{graphicx}

\begin{document}

\maketitle

\begin{abstract}

Recent advancements in multimodal foundation models have showcased impressive capabilities in understanding and reasoning with visual and textual information. 
Adapting these foundation models trained for general usage to specialized domains like biomedicine requires large-scale domain-specific instruction datasets.
While existing works have explored curating such datasets automatically, the resultant datasets are not explicitly aligned with domain expertise.
In this work, we propose a data-centric framework, \textbf{Biomed}ical \textbf{V}isual \textbf{I}nstruction \textbf{T}uning with Clinician Preference \textbf{Al}ignment (\ours), that incorporates clinician preferences into both stages of generating and selecting instruction data for tuning biomedical multimodal foundation models. 
First, during the generation stage, we prompt the GPT-4V generator with a diverse set of clinician-selected demonstrations for preference-aligned data candidate generation. Then, during the selection phase, we train a separate selection model, which explicitly distills clinician and policy-guided model preferences into a rating function to select high-quality data for medical instruction tuning. 
Results show that the model tuned with the instruction-following data from our method demonstrates a significant improvement in open visual chat (18.5\% relatively) and medical VQA (win rate up to 81.73\%). Our instruction-following data and models are available at \url{https://BioMed-VITAL.github.io}.

\end{abstract}

\section{Introduction}
\label{sec:intro}
Recent advancements in large pre-trained multimodal models, such as GPT-4V~\cite{achiam2023gpt}, have demonstrated impressive performance on various language and vision tasks. 
However, when directly applied to specialized domains like biomedicine, these models may fall short due to their primary focus on general usage rather than domain-specific expertise~\cite{wu2024pmc,doi:10.7326/M23-2772}. 
To bridge this gap and adapt general domain models to specialized domains, researchers have explored various techniques. Instruction tuning has emerged as a promising approach, involving the fine-tuning of large foundation models to follow explicit, natural language instructions~\cite{DBLP:conf/iclr/WeiBZGYLDDL22,DBLP:conf/acl/MishraKBH22,zhang2023instruction}. These instructions are composed of task-specific prompts and their corresponding response, enabling the models to learn and generalize to a wide range of tasks within the target domain. 

Although instruction tuning has proven to be an effective method for adapting models to target domains and performing various downstream tasks, its success heavily depends on large-scale instruction-following datasets. Curating large-scale instructional datasets in specialized domains, such as biomedicine, can be expensive and time-consuming, often requiring significant domain expertise. Previous work proposes to use strong language models to generate instruction data automatically, which effectively reduces the need for extensive manual annotation~\cite{alpaca}.
Such paradigms have successfully been adopted to adapt general domain models to biomedicine. For example, LLaVA-Med~\cite{li2024llava} developed a framework to instruction-tune biomedical language-vision models with GPT-4 generated instruction-following data. This approach has achieved impressive performance on open-ended visual chat and visual question answering benchmarks, highlighting the potential of using model-generation data in the biomedical domain.

However, existing methods for automatically curating datasets do not explicitly incorporate clinician preferences, which may result in models producing irrelevant or impractical output, limiting their utility in real-world applications~\cite{fleming2024medalign}. 
Yet, aligning domain expertise with the process of instruction-following datasets curation is challenging. 
First, advanced data generators, such as GPT-4V, are often proprietary and not publicly available for alignment tuning. 
Second, clinician-annotated preference data in the biomedical domain is limited, further restricting effective preference learning. 
The combination of model opacity and data scarcity creates a significant bottleneck in developing high-quality, expert-aligned instruction-following data for instruction-tuning. This hinders the development of domain-specific models that can effectively incorporate expert preferences and requirements, ultimately limiting their practical utility and real-world impact.

To tackle this challenge, we propose an effective data-centric approach, \ours, that incorporates clinician preference into the process of automatically curating instruction-following data for biomedical visual instruction tuning.
As shown in Figure~\ref{fig:data-gen-framework}, \ours consists of three stages: (1) data generation with demonstrations, (2) data selection with a preference distilled model, and (3) visual instruction-tuning.
In \underline{data generation}, we strategically sample a diverse set of instructions to collect clinician preferences, which are used as demonstrations for GPT-4V-based instructional data generation, guiding the data generation toward producing more clinically relevant and useful instruction-following examples.
In the \underline{data selection} stage, we train a data selection model that distills a mixture of preferences from clinician-annotated and model-annotated data guided by clinician-curated criteria. This model is then used to rank the generated data samples, and the top-ranked samples are selected for visual instruction-tuning.

The contributions of this work are summarized as follows:
\begin{itemize}[nosep,leftmargin=*]
\item  We introduce a data-centric framework \ours, which generates and selects instruction-following data aligned with clinician preference for visual instruction tuning. Evaluation indicates an improved data quality and our instruction-tuned models remarkably improve in both open visual chat (18.5\% relatively) and three biomedical VQA benchmarks (win rate up to 81.73\%). 
\item We propose a paradigm involving clinician preference during generation and an effective data selection model based on a mixture of preferences. It is shown that our distilled data selection model excels in matching human preferences compared with judgments of GPT-4. 
\item To facilitate further study, we release 80K clinician preference-aligned instruction-following datasets generated and selected from ours, along with the models instruction-tuned based on them. All resources are publicly available on the website \url{https://BioMed-VITAL.github.io}.
\end{itemize}

\section{Background}
\label{sec:background}

\textbf{Instruction-Tuning.}
Instruction tuning has become an effective method for adapting pre-trained language models to a wide range of natural language tasks~\cite{zhao2021calibrate, wang2023selfinstruct, wan2023gptre, yao2023tree, gao2023enabling, ning2023album, sun2023text, chen2024selfplay} by providing task-specific instructions and examples. This approach has been further explored in studies like FLAN-T5~\cite{chung2022scaling}, LLaMA~\cite{touvron2023llama}, and LLaMA2~\cite{touvron2023llama2}, which enables models to understand and follow task-specific instructions without extensive task-specific fine-tuning. Recently, using strong language models to generate instruction data automatically has been proposed to train a high-quality instruction-following model under an academic budget~\cite{peng2023instruction, alpaca, liu2023visual}. For example, Stanford Alpaca~\cite{alpaca} instruction-tuned LLaMA using \texttt{text-davinci-003}-generated instruction-following datasets and achieved competitive performance on various NLP tasks. 

\vspace{-5pt}

\textbf{Vision-Language Foundation Models in Biomedical Domain.}
General vision-language foundation models have achieved remarkable success across various domains. Researchers in biomedicine have been actively exploring the adaptation of vision-language foundation models to tackle domain-specific tasks~\cite{moor2023foundation,Bedi2024.04.15.24305869,plip,saab2024capabilities}.
However, effectively adapting vision-language foundation models to specialized domains such as the biomedical presents challenges, particularly due to limited training data. 
To overcome this challenge, our work aims to establish a data-centric method that aligns domain expertise from clinicians with the instructional data for instruction-tuning, which generates and selects instruction-following datasets that are aligned with clinician preference. 
\section{Clinician-Aligned Biomedical Visual Instruction Tuning}

\begin{figure}[t]
\centering
  \includegraphics[width=\textwidth]{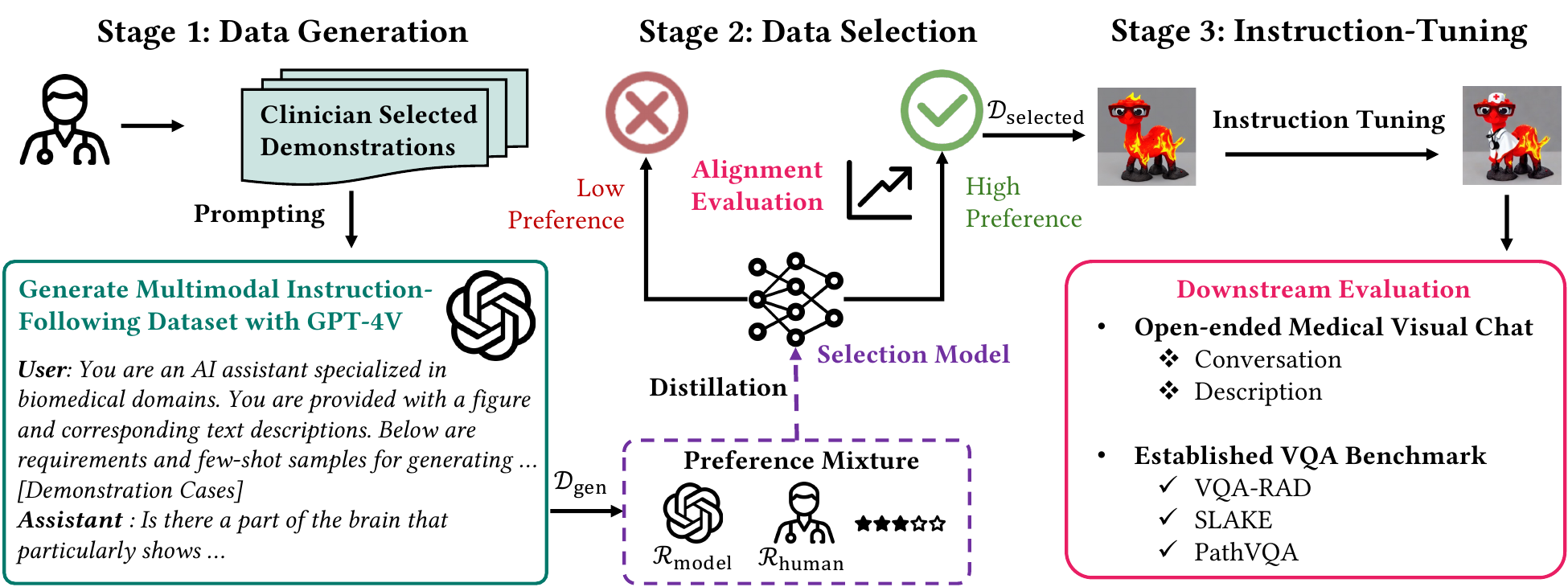}
    \caption{Overview of \textbf{Biomed}ical \textbf{V}isual \textbf{I}nstruction \textbf{T}uning with Clinician Preference \textbf{Al}ignment (\ours). Clinician preferences are infused in the 1. data generation and 2. selection stages.}
    \label{fig:data-gen-framework}
\end{figure}

Figure~\ref{fig:data-gen-framework} presents an overview of the proposed framework \ours, consisting of three stages: (1) data generation with diverse expert-selected demonstration, (2) data selection with a distilled selection model trained with mixed preferences, and (3) instruction tuning to adapt a general multimodal model for biomedical tasks. 
The output from the framework includes a clinician preference-aligned instruction-following dataset $\mathcal{D}=\{(I_{i}, C_{i}, \mathbf{Q}_{i}, \mathbf{A}_{i})\}_{i=1}^N$ and instruction-tuned models based on it. 
$I_{i}$ represents the $i$-th biomedical image; $C_{i}$ is the caption and inline-mentions associated with the $i$-th image; $\mathbf{Q}_{i}=\left\{\mathcal{Q}_{ij}\right\}_{j=1}^{n_i}$ contains $n_i$ instructions, where $j$ represents the $j$-th instruction for the $i$-th image-text sample; $\mathbf{A}_{i}=\left\{\mathcal{A}_{ij}\right\}_{j=1}^{n_i}$ contains $n_i$ responses, each corresponding to $\mathcal{Q}_{ij}$; and $N$ is the total number of samples in the dataset. 

\subsection{Stage 1: Data Generation with Diverse Expert-Selected Demonstration}
Large pre-trained models have shown strong in-context learning capabilities by learning from a few presented examples and mimicking when generating responses. In \ours, we use the GPT-4V model as the generator. To incorporate clinician preference into the data generation process, we first select a diverse set of samples for clinicians to annotate. Clinician-selected QA pairs are used as few-shot demonstrations for GPT-4V to generate instruction-following data at scale.

\textbf{Diverse few-shot demonstration selection.}
We employ a strategic sampling approach to ensure the diversity and representatives of the demonstrations for the generator. For each sample $(I_{i}, C_{i})$ in the dataset, the image and text representations are extracted using BiomedCLIP~\cite{zhang2024biomedclip}, then K-means clustering is performed on these representations to cluster the samples into K distinct categories, denoted as ${\mathcal{D}_1, \mathcal{D}_2, ..., \mathcal{D}_\text{K}}$. From these clusters, we uniformly select a subset $S = {(I_{i}, C_{i})}_{i=1}^M$ with total $M$ samples that have relatively complex captions and inline mentions. For each selected sample $(I_{i}, C_{i}) \in S$, we use GPT-4V to generate a set of instructions $\mathbf{Q}_i=\left\{\mathcal{Q}_{ij}\right\}_{j=1}^{n_i}$, and two candidate responses ${A_{ij}^\text{1}, A_{ij}^\text{2}}$ for each instruction $\mathcal{Q}_{ij}$.
Clinicians are presented with these generated responses and are asked to choose the better one $A_{ij}^\text{pref}$ between the two candidates, select both if two responses are equally good, or deselect both to drop this instruction. The resulting annotation $\mathcal{R}_\text{human}$ contains the selected preferences from clinicians. 


\textbf{Instruction-following data generation with GPT-4V.}
Using the clinician-selected data, we employ GPT-4V as the generator to simulate the instructional dataset. During each API call, we randomly select 2 samples for each of the 5 modalities from $\mathcal{D}_\text{pref}$ as few-shot demonstrations and append them to the language prompts. 
The full prompt can be referred to in Appendix~\ref{app:data_generation_prompt}. 
Compared with previous methods, our generated dataset $\mathcal{D}_\text{gen} = \left\{(I_i, C_i, \mathbf{Q}_i, \mathbf{A}_i)\right\}_{i=1}^N$ incorporate visual input and is further guided with selected clinician demonstrations. 


\subsection{Stage 2: Distilling Mixed Clinician Preference for Data Selection}
While $\mathcal{D}_\text{gen}$ is directly usable to instruction-tune, it may still include samples that can introduce noise or bias or are irrelevant to the real needs of clinicians. 
In the second stage of \ours, we train a data selection model that learns to select instruction data aligned with expert preference. 


\textbf{Preference data from two resources.}
Collecting human preference data from domain experts such as clinicians is expensive and time-consuming. Thus, the available annotation data is usually on a small scale. A recent paradigm involves using LLMs as judges, which have been shown to match human preferences effectively~\cite{DBLP:conf/nips/ZhengC00WZL0LXZ23}.
We consider a data mixing schema to distill preference into a local model for data selection. Our preference data comes from two resources, from humans and from models: 
(1) human preference from the preference annotation $\mathcal{R}_{\text{human}}$ in stage 1, where each question $\mathcal{Q}_{ij} $ is paired with two candidate answers ${A_{ij}^{1}, A_{ij}^{2}}$, with $A_{ij}^\text{pref}$ annotated as the preferred one. 
(2) model-based preference: to generate reliable model-based ratings, we first collect a set of clinician-curated factors for data quality evaluation, such as missing information, recognition errors, lack of medical precision, insufficient depth, valueless questions, etc. With these clinician-curated criteria, we use GPT-4V as a judge to score a randomly sampled set of data from 0 to 10. The detailed prompt can be referred to in Appendix~\ref{app:model_preference_prompt} Figure~\ref{fig:model-preference}. The resulting self-evaluated ratings, $\mathcal{R}_{\text{model}}$, provide additional preference data and address the scalability issue related to human annotation.

\textbf{Distill clinician preference to a selection model.}
Next, we train a data selection model with the preference data, which is designed to identify and remove low-quality samples from the generated dataset and preserve only the most accurate and clinically relevant examples for instruction tuning. 
We use BiomedCLIP~\cite{zhang2024biomedclip} as the backbone, followed by an MLP head to perform binary prediction tasks on good/bad ratings of data samples. Pairwise ranking loss is used as the training objective: given a pair of candidate samples $x_i$ and $x_j$, along with their corresponding annotated preferences $\mathcal{R}_i$ and $\mathcal{R}_j$, the objective is formulated as a pairwise classification:
\begin{equation}
\mathcal{L}_Q=-z_i \log \sigma\left(f(x_i)\right)-z_j \log \sigma\left(f(x_j)\right),
\label{eq:loss}
\end{equation}
where $\sigma$ represents the sigmoid function, and $f(\cdot)$ denotes the rating function learned by the model. The values of $z_i$ and $z_j$ are determined by comparing the preference annotation:
\begin{equation}
\left(z_i, z_j\right)=\left\{\begin{array}{ll}
(1,0), & \mathcal{R}_i \geq \mathcal{R}_j \\
(0,1), & \mathcal{R}_i < \mathcal{R}_j
\end{array} .\right.
\label{eq:z-value}
\end{equation}
By minimizing the pairwise classification loss, the data selection model learns to assign higher scores to samples with higher preference and lower scores to samples with lower preference. 

\textbf{Preference mixing strategy during training.} We mix two sources of preference data in each batch during training. In Eq~(\ref{eq:loss}), each $x_i$ and $x_j$ can be either human-annotated preferences from $\mathcal{R}_{\text{human}}$, or two samples with model-based ratings $\mathcal{R}_{i}$ and $\mathcal{R}_{j}$ from $\mathcal{R}_{\text{model}}$. To address the scalability difference between the two resources, we introduce an adaptive contribution mechanism by incorporating an adjustable sample weight $w$ into Eq~(\ref{eq:loss}):
\begin{equation}
\mathcal{L}_Q=-w \left(z_i \log \sigma\left(f(x_i)\right)+z_j \log \sigma\left(f(x_j)\right)\right),
\label{eq:adaptive-loss-weighted}
\end{equation}
where $w$ allows for an adjustable contribution of the two preference data resources during training.

\textbf{Data selection with distilled selection model.}
We apply the trained data selection model to the generated dataset $\mathcal{D}_\text{gen}$ and observe F1@\texttt{K} and Precision@\texttt{K} curves to determine the threshold for data selection. To balance data quality and diversity, we first cluster all the data samples into K groups and uniformly select top-ranked data in each group to compose the final instruction-following dataset, denoted as $\mathcal{D}_\text{selected}$, which contains the most informative, accurate, and clinically relevant examples. More empirical decisions during selection are discussed in Section~\ref{sec:intermediate}. 

\vspace{-0.5em}
\subsection{Stage 3: Instruction-Tuning}
Following LLaVA-Med~\cite{li2024llava}, we continue training the LLaVA~\cite{liu2023llava,liu2023improvedllava} model on our curated instruction-following dataset $\mathcal{D}_\text{selected}$.
The instruction tuning objective for model $\theta$ is to minimize the negative log-likelihood of the target $\mathbf{A}_i$ given input image $I_i$, caption $C_i$, question $\mathbf{Q}_i$,
\begin{equation}
\mathcal{L}_{IT} = -\sum_{i=1}^{|\mathcal{D}_\text{selected}|} \log p(\mathbf{A}_i | I_i, C_i, \mathbf{Q}_i, \theta).
\end{equation}

\vspace{-10pt}
\vspace{-0.5em}
\section{Experiments}
\subsection{Dataset and Experiment Details of \ours}
We follow the setup of Li et al.~\cite{zhang2024biomedclip} and utilize image-text pairs from the PMC-15M dataset~\cite{zhang2024biomedclip} to generate multi-round QA instructional data. For the data generator, we utilize \texttt{gpt-4-vision-preview} API on Azure OpenAI. For the data selector, we use BiomedCLIP~\cite{zhang2024biomedclip}, which is trained for 6 epochs with a learning rate of 1e-4. For instruction-tuning, we use \texttt{llava-v1.5-13b} as the backbone. Following the LLaVA-Med approach~\cite{li2024llava}, the model is first trained with biomedical concept alignment; subsequently, it is instruction-tuned using the selected dataset from the second stage, utilizing a multi-turn dialogue setup~\cite{liu2023visual}. The instruction-tuning process is carried out for 3 epochs with a learning rate of 2e-5, trained and tested with 2 NVIDIA A100 GPUs.


\vskip -0.5em
\subsection{Alignment Evaluation of the Data Selection Model}
\label{sec:intermediate}

\begin{wraptable}{l}{5.5cm}
\centering
\vskip -0.5em
\caption{Rank-based metrics by varying data mixture weight $w$.}
\resizebox{0.85\linewidth}{!}{
\begin{tabular}{lcccc}
\toprule
\multirow{2.5}{*}{$w$} & \multicolumn{4}{c}{Rank-based Metrics (\%)}                  \\ 
\cmidrule(l){2-5} 
& ACC $\uparrow$ & AUC $\uparrow$ & MR $\downarrow$ & \multicolumn{1}{c}{MAP} $\uparrow$ \\ 
\midrule
1 & 61.61 & 61.90 & 44.04 & 62.22 \\
5     & 58.63    & 58.22    & 45.87    & 62.29 \\
10    & 59.38    & 59.14    & 45.39    & 59.20  \\
50    & 59.67          & 59.80           & 45.09          & 63.27 \\
100   & 62.05          & 62.30           & 43.84          & 61.63 \\
200   & 60.91          & 61.23          & 44.37          & 59.55 \\
300   & 63.64          & 63.12          & 43.43          & 63.00 \\
\textbf{400}  & \textbf{66.72} & \textbf{66.32} & \textbf{41.83} & \textbf{64.47} \\
500   & 62.85          & 63.06          & 43.46          & 65.00 \\
600   & 56.30           & 56.07          & 46.95          & 60.25 \\ 
\bottomrule
\end{tabular}
}
\label{table:intermediate}
\end{wraptable}

\textbf{Preference mixture.} 
In our experiment, we train the selection model with a proportion of 1:400 of human and model preferences to reflect the scalability gap. To find the optimal balance between the two resources, we adjust the adaptive factor $w$ in Eq~(\ref{eq:adaptive-loss-weighted}) and observe the selection model performance trained with the data mixture under $w$ on a clinician-annotated test set.
The results in Table~\ref{table:intermediate} show that the best performance is achieved when $w$ is 400, which happens to balance the contribution of human and model-annotated preference in the total training loss. This finding suggests that with the data mixture, both resources are beneficial and complementary for learning preference.
The results also reveal a successful distillation of clinician preferences into the selection model. 

\vskip -1em
\begin{figure}[h]
\includegraphics[width=\linewidth]{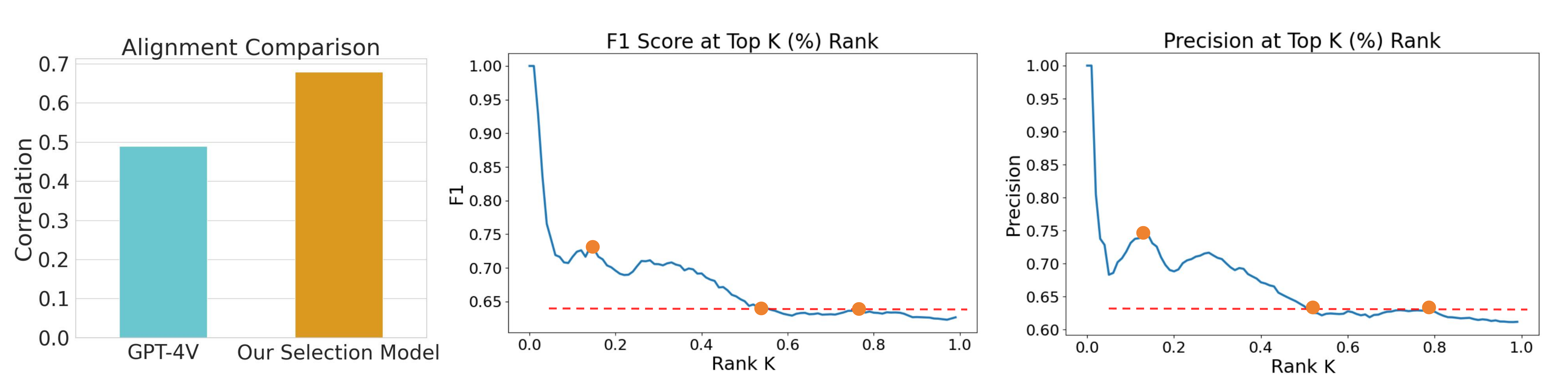} 
\caption{Left: Comparison of human preference alignment between GPT-4V and our selection model. Right: F1 and precision for varying top \texttt{K} percentile samples ranked by the selection model.}
\label{fig:eval-selection}
\end{figure}

\textbf{Alignment with human preference versus GPT-4.}
To compare the preference evaluation ability of our trained data selection model versus the GPT-4 model, we calculate the correlation between the ratings generated from both models with gold clinician-annotated preference. The results in Figure~\ref{fig:eval-selection} (left panel) indicate a better alignment of our trained selection model over the GPT-4 model.

\textbf{Selecting top \texttt{K} ranked samples.} We observe the F1 
and Precision performance curves on the ranking list from the score model by varying the top \texttt{K} percentiles to determine the optimal proportion of top-ranked data to select. As illustrated by Figure~\ref{fig:eval-selection} (right panel), we identify three critical percentiles: 10\%, 50\%, and 80\%, where the performance either reaches a local peak or plateaus afterward, indicating that further incorporating data on the ranking list would not yield significant improvements. Consequently, we select datasets corresponding to these critical percentiles for visual instruction tuning, ensuring that the models learn from high-quality, clinician-preferred samples.


\subsection{Downstream Evaluation 1: Open-Ended Medical Visual Chat}
To evaluate the model's ability to engage in dialogue-like interactions and provide coherent responses, we evaluate the model with open-ended visual chat, where the trained language models are prompted to respond to given questions based on the provided images and texts in a multi-round manner.

\textbf{Dataset and evaluation paradigm.}
For the evaluation dataset, we use 50 unseen image and caption pairs with 193 question-answer pairs collected by the LLaVA-Med~\cite{li2024llava} authors. 
The questions are divided into two types:
(1) Conversation questions, which require the model to engage in dialogue-like interaction, understand the context and provide relevant responses. For example, given an image of a chest X-ray, a conversation question might ask, ``What abnormalities do you see in this X-ray image?''
(2) Description questions, which focus on detailed descriptions or explanations based on visual and textual input. For instance, a description question for a histology image could be, ``Describe the morphological features of the cells in this histology slide.''

To evaluate the quality of the model's open-ended responses, we use GPT-4V as the evaluator. A reference prediction is first generated based on the input context and the given question, which is then provided to assess the responses from various trained models by assigning a relative score on a scale from 1 to 10. A higher score indicates that the model's response is more accurate, relevant, and coherent with respect to the reference prediction. 

\begin{table*}[h]
\begin{center}
\caption{Performance comparison of the instruction-tuned models on open-ended biomedical visual chat. The number followed by ``\#: '' represents the number of testing samples in this category. In the following experiments, $N$ is the number of QA pairs of 60K images. }
\resizebox{\linewidth}{!}{
\begin{tabular}{lccccccccccc}
\toprule
\multirow{3.5}{*}{\bfseries Model} & \multirow{3.5}{*}{\bfseries Data Size} & \multicolumn{2}{c}{\bfseries Question Types} && \multicolumn{5}{c}{\bfseries Domains} && \multirow{2.5}{*} {\bfseries Overall}   \\ 
\cmidrule(lr){3-4} \cmidrule(lr){6-10} 
& & Conversation & Description && CXR & MRI & Histology & Gross & CT &&   \\
& & (\#:143) & (\#: 50) && (\#: 37) & (\#: 38) & (\#: 44) & (\#: 34) & (\#: 40) && (\#: 193)  \\
\midrule
LLaVA-Med & $N$ & 58.53 & 56.16 && 43.97 & 51.19 & 60.01 & 86.49 & 50.63 && 57.92 \\
\ours & Top 10\% $*N$ & 64.11 & 60.05 && 56.35 & 52.57 & 59.02 & 87.60 & 62.82 && 63.06 \\
\ours & Top 50\% $*N$ & 65.95 & 64.26 && 55.75 & 55.57 & 60.96 & 94.06 & 64.70 && 65.51 \\
\ours & Top 80\% $*N$ & 68.50 & \textbf{67.65} && 55.24 & \textbf{58.73} & 62.65 & \textbf{101.88} & \textbf{67.05} && 68.28 \\
\ours & $N$ & \textbf{69.73} & 65.51 && \textbf{59.22} & 57.39 & \textbf{67.15} & 99.26 & 63.63 && \textbf{68.63} \\
\midrule
\rowcolor{gray!15}\multicolumn{12}{l}{\emph{Model Ablation}}  \\
\oursv$^\text{A0}$ & $N$ & 65.38 & 60.63 && 63.48 & 53.82 & 57.32 & 92.30 & 58.16 && 64.15 \\
\oursv$^\text{A1}$ & $N$ & 67.82 & 59.48 && 59.68 & 53.98 & 60.34 & 97.89 & 60.74 && 65.66 \\
\oursv$^\text{A2}$ & $N$ & 67.53 & 62.78 && 60.64 & 54.62 & 61.07 & 98.27 &  61.21 && 66.30 \\
\bottomrule
\end{tabular}
}
\label{tab:sec7-conversation}
\end{center}
\end{table*}

\textbf{Model variants.}
In addition to comparing our model with the LLaVA-Med baseline, we further investigate the influence of the selected data size on instruction tuning performance and conduct a model ablation study. \ding{169} To study the impact of data size, we instruction-tune three additional models using datasets selected from the ranking list at three critical percentiles: 10\%, 50\%, and 80\%, as described in Section~\ref{sec:intermediate} and illustrated in Figure~\ref{fig:eval-selection}. \ding{169} For the model ablation study, we include three variants based on the full \ours model: \oursv$^\text{A0}$, which does not incorporate clinician preference alignment in either stage; \oursv$^\text{A1}$, which only includes the first stage of clinician-selected demonstrations; and \oursv$^\text{A1}$, which only incorporates the second stage of preference distillation. The results of these investigations are summarized in Table~\ref{tab:sec7-conversation}.

\textbf{Result discussion.} For the three-dimensional comparison: 
\begin{itemize}[nosep,leftmargin=*]
\item \underline{Baseline comparison}: \ours and all its variants consistently outperform the compared method. Even with only the top 10\% of selected data, the \ours model surpasses the baseline model trained on the full dataset of size $N$ in both question types, highlighting the effectiveness of our data-centric framework.
\item \underline{Data size study}: When varying the top-ranked percentiles in the data selection process, increasing the dataset size generally improves model performance. Notably, our models trained with fewer data (i.e., 50\% and 80\% of the dataset) outperform the \oursv$^\text{A0}$ and \oursv$^\text{A1}$ models, which are trained on the full data size $N$ without data selection. This finding suggests that the second-stage data selection leads to more efficient and effective model tuning, as it focuses on the most informative and relevant examples.
\item \underline{Model ablation study}: Comparing the three model ablations with the full \ours model, we observe that incorporating clinician preference infusion in both the data generation and selection stages leads to improved performance compared to the base model. The full \ours model achieves the best performance, revealing the effectiveness of combining both alignment stages to achieve optimal results. This finding underscores the importance of considering clinician preferences throughout the entire data-centric framework for biomedical visual instruction tuning.
\end{itemize}

\subsection{Downstream Evaluation 2: Performance on Established VQA Benchmarks}

\textbf{Dataset details.}
We train and evaluate \ours on three widely used biomedical visual question answering benchmarks~\cite{li2024llava, tu2024towards, zhang2023pmcvqa}. 
The statistics of the datasets are shown in Table~\ref{tab:vqa-stats}. 

\begin{table}
\centering
\caption{Statistics of the benchmark datasets for downstream evaluation on biomedical VQA.}
\resizebox{0.75\columnwidth}{!}{%
\begin{tabular}{lcccccccccc}
\toprule
\multirow{2.5}{*}{Dataset}  & \multicolumn{2}{c}{\bf VQA-RAD} & & \multicolumn{3}{c}{\bf SLAKE} & & \multicolumn{3}{c}{\bf PathVQA}\\
\cmidrule(lr){2-3} \cmidrule(lr){5-7} \cmidrule(lr){9-11} 
& Train & Test && Train & Val & Test && Train & Val & Test  \\
\midrule
\# Images & 313 & 203 && 450 & 96 & 96 && 2,599 & 858 & 858 \\
\# QA Pairs & 1,797 & 451 && 4,919 & 1,053 & 1,061 && 19,755 & 6,279 & 6,761 \\
\# Open & 770 & 179 && 2,976 & 631 & 645 && 9,949 & 3,144 & 3,370 \\
\# Closed & 1,027 & 272 && 1,943 & 422 & 416 && 9,806 & 3,135 & 3,391 \\
\bottomrule
\end{tabular}
}
\label{tab:vqa-stats}
\end{table}

\begin{itemize}[nosep,leftmargin=*]
\item VQA-RAD~\cite{lau2018dataset} is a dataset containing 3,515 question-answer pairs created by medical professionals, along with 315 radiology images. Each image is linked to several questions, which are categorized into 11 types, including abnormality, attribute, modality, organ system, color, counting, object/condition presence, size, plane, positional reasoning, and others. The dataset features a balanced mix of closed-ended (yes/no) and open-ended (one-word or short phrase) answers.
\item SLAKE~\cite{liu2021slake} is a comprehensive medical visual question-answering dataset with knowledge-enhancement features. It contains radiology images and diverse question-answer pairs annotated by experienced physicians. The dataset incorporates external medical knowledge through a provided medical knowledge graph, and the images are supplemented with rich visual annotations, including semantic segmentation masks and object detection bounding boxes. SLAKE covers a wide range of modalities and human body parts, such as the brain, neck, chest, abdomen, and pelvic cavity. We adopt only the English subset of SLAKE in our experiments.
\item PathVQA~\cite{he2020pathvqa} focuses on pathology images. Each image is associated with multiple questions that cover various aspects, such as location, shape, color, and appearance. The questions in PathVQA include open-ended questions (e.g., why, what, how, where) and closed-ended questions. 
\end{itemize}

\textbf{Experimental details.}
For each benchmark, the model is fine-tuned for 15 epochs with a learning rate of 2e-5. To account for the open-ended nature and expressive diversity of language generation, we report both metrics-based performance and an additional model-based win rate performance. The win rate performance provides a complementary perspective on the model's ability to generate accurate and relevant responses compared to the baseline.

\begin{table*}[h]
\begin{center}
\caption{Metric performance of \ours and compared methods on three VQA benchmarks.  Models based on LLaVA are trained with 7b/13b backbone and training sample size of 60K/150K. The largest set 150K combines 10K and 60K provided by LLaVA-Med, plus our curated 80K samples. }
\resizebox{\linewidth}{!}{
\begin{tabular}{lcccccccccccc}
\toprule
\multirow{2.5}{*}{\bfseries Model} && \multicolumn{3}{c}{\bfseries VQA-RAD} & & \multicolumn{3}{c}{\bfseries SLAKE} & & \multicolumn{3}{c}{\bfseries PathVQA} \\ 
\cmidrule(lr){3-5} \cmidrule(lr){7-9} \cmidrule(lr){11-13} 
&& Ref & Open & Closed && Ref & Open & Closed && Ref & Open & Closed \\
\midrule
\rowcolor{gray!15}\multicolumn{13}{l}{\emph{Supervised fine-tuning results from models based on LLaVA (model size, training sample size)}}  \\
LLaVA (7b, 60K) && & 50.00 & 65.07 & && 78.18 & 63.22 & && 7.74 & 63.20 \\
LLaVA-Med (7b, 60K) && & 61.52 & 84.19 & && 83.08 & 85.34 & && 37.95 & 91.21 \\
LLaVA-Med (13b, 60K) && & 64.58 & 77.94 & && 84.97 & 85.58 & && 38.82 & \underline{92.39} \\
\ours (13b, 60K) & && \underline{64.88} & \underline{84.55} & && \underline{87.82} & \underline{86.54} & && \underline{39.71} & 91.41 \\
\ours (13b, 150K) & && \textbf{69.72} & \textbf{84.86} & && \textbf{91.69} & \textbf{90.70} & && \textbf{39.89} & \textbf{92.42} \\
\midrule
\rowcolor{gray!15}\multicolumn{13}{l}{\emph{Literature-reported results from representative SoTA methods}}  \\
MMQ~\cite{do2021multiple} && 53.70 &  & 75.80 &  &  &  &  & &  13.40 & & 84.00 \\
Prefix T. Medical LM~\cite{van2023open} && & &  & & 84.30 & & 82.01 & & 40.00 & & 87.00 \\
PubMedCLIP~\cite{eslami2023pubmedclip} && 60.10 &  & 80.00 &  & 78.40 &  & 82.50 & & & & \\
BiomedCLIP~\cite{zhang2023large} && 67.60 &  & 79.80 &  & 82.05 &  & 89.70 & & & & \\
M2I2~\cite{li2023self} && 66.50 &  & 83.50 &  & 74.70 &  & 91.10 & & 36.30 & & 88.00 \\
MUMC~\cite{li2023masked} && 71.50 &  & 84.20 &  & 81.50 &  & 81.50 & & 39.00 & & 65.10 \\
M3AE~\cite{chen2022multi} && 67.23 &  & 83.46 &  & 80.31 &  & 87.82 & &  & &  \\
CoQAH~\cite{kim2024generalizing} && 30.20  &  & 67.50 &  & 42.50  &  & 73.90  & &  & &  \\
PMC-CLIP~\cite{lin2023pmc} && 67.00 &  & 84.00 &  & 81.90 &  & 88.00 & &  & &  \\
\bottomrule
\end{tabular}
}
\label{tab:vqa}
\end{center}
\end{table*}

\textbf{Metric performance.}
To evaluate the performance metrics, we follow the practice of Li et al.~\cite{li2024llava} and use accuracy for closed-set questions and recall (the ratio of ground-truth tokens appearing in the generated response) for open-set questions. 
Table~\ref{tab:vqa} summarizes the metric performance of \ours compared to models based on LLaVA, as well as literature-reported results from representative state-of-the-art (SoTA) methods for reference\footnote{Details of the compared SoTA methods can be referred to in Appendix~\ref{app:baseline}.}. 
Among the supervised fine-tuning models based on LLaVA, \ours consistently outperforms the other two, particularly on open-type questions, with the 150K trained model achieving the best. 
When comparing ours to those reported in the literature from previous methods, it is important to note that some prior methods formulate the problems as classification tasks among answer candidates in the training set, which does not meet the real-world need for open-ended QA. Additionally, some studies report metrics on the open set using different calculations, leading to inconsistencies in comparison. We follow the practice of Li et al.~\cite{li2024llava} and present the numbers from prior work only as a reference for the open set while including metrics on the closed set for comparison. The results demonstrate that \ours achieves leading performance in most cases, even when compared to methods that employ classification set up for QA despite \ours being in an open, generative manner.


\begin{figure}
\includegraphics[width=\linewidth]{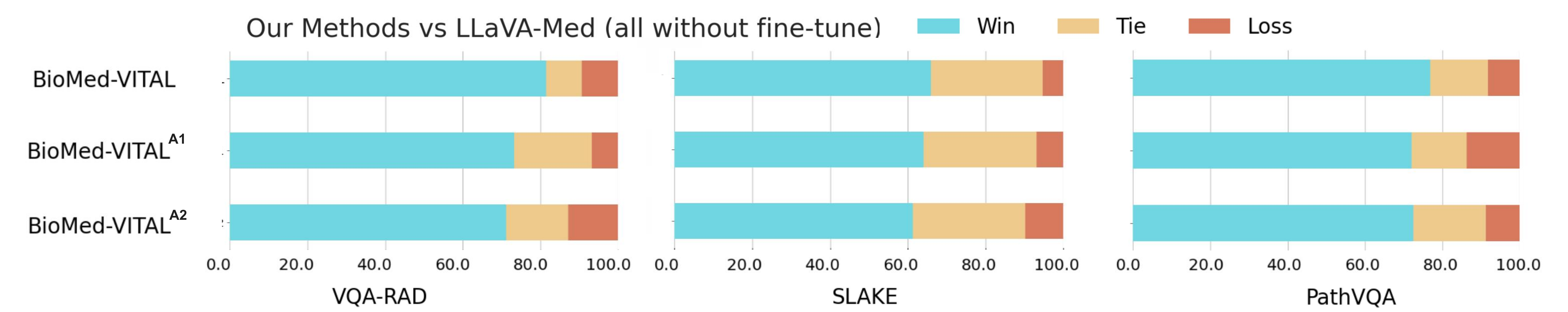} 
\caption{Win rate performance of \ours and its variants compared with LLaVA-Med.}
\label{fig:winrate}
\end{figure}

\textbf{Win rate performance.}
Recent studies in visual question-answering have highlighted the limitations of token-matching metric evaluation for open-ended language generation tasks and have proposed leveraging model-based win rate evaluation instead~\cite{Manas_Krojer_Agrawal_2024,hu2023tifa}. In line with these insights, we adopted a reference-guided win rate evaluation, where GPT-4V is employed as an impartial judge to assess the quality of the responses provided by two compared models. 
The detailed prompt for win rate evaluation on VQA benchmarks is shown in Appendix~\ref{app:evaluation_prompt} Figure~\ref{fig:win-rate}. 
By considering the ground-truth reference, GPT-4V determines which model provides the more accurate and relevant answer, offering a comprehensive evaluation of the models' performance in reponse generation. 

As shown in Figure~\ref{fig:winrate}, \ours and its variants \oursv$^\text{A1}$ and \oursv$^\text{A2}$ outperform the LLaVA-Med baseline and achieve significantly higher win rates up to 81.73\%. It is worth noting that the full model consistently performs the best compared to the two ablations, indicating the effectiveness of the clinician preference alignment during both the data generation and selection phases. Between the two model variants, \oursv$^\text{A1}$, which only incorporates clinician alignment in the data generation phase, performs slightly better than \oursv$^\text{A2}$, which only incorporates clinician alignment in the data selection phase. This finding indicates the greater impact of the generation phase on clinician preference alignment than the selection phase.

\subsection{Case Study}
\label{sec:case}


\definecolor{mygreen}{RGB}{48,192,180}
\definecolor{myblue}{RGB}{72,116,203}
\definecolor{myred}{RGB}{229,76,94}

\begin{figure}[h]
\includegraphics[width=\linewidth]{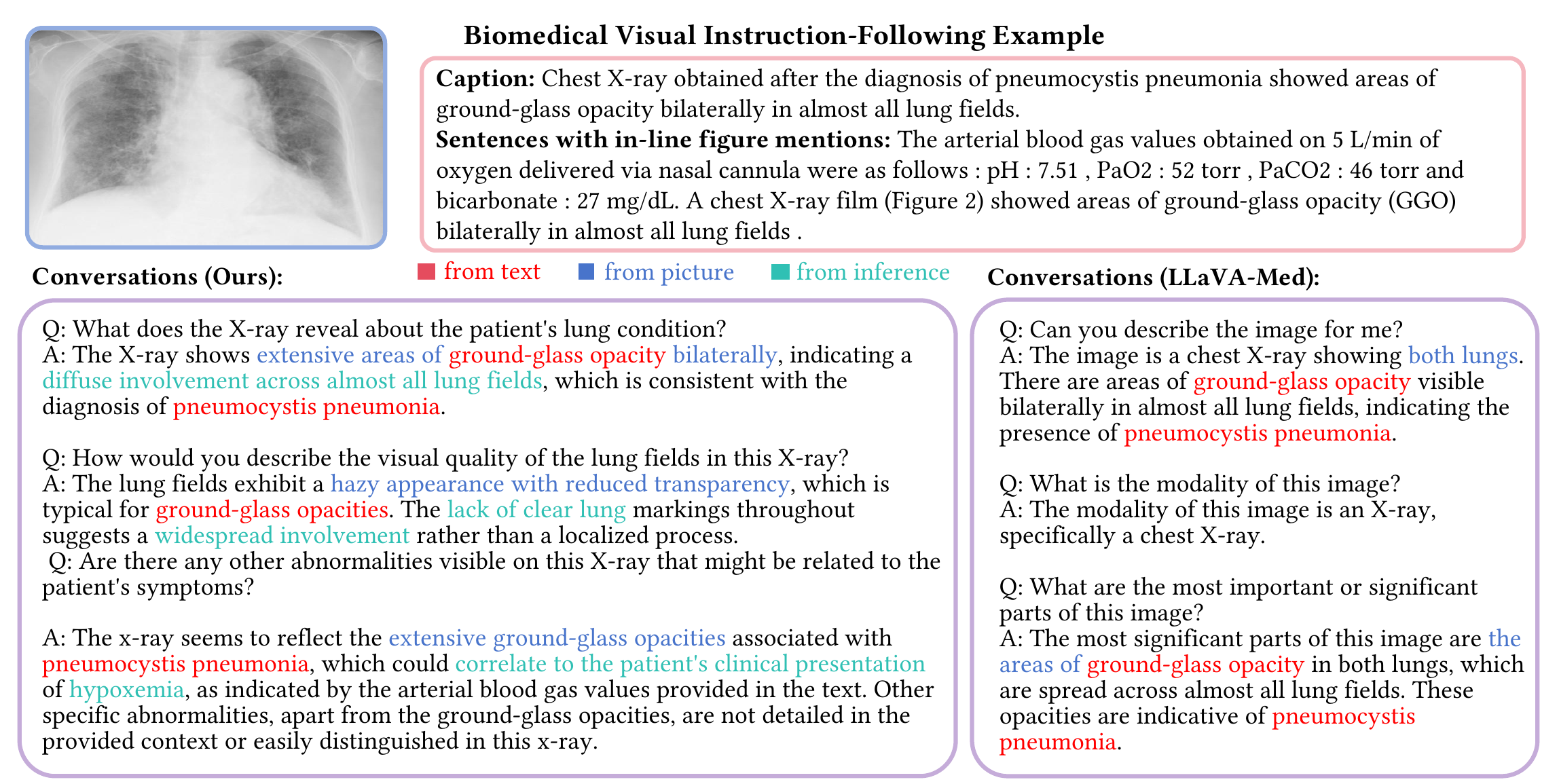} 
\caption{Case study on the generated instruction-following data.}
\label{fig:instruct-following}
\end{figure}

\textbf{Generated instruction-following data.}
We present case studies of the instructional data produced by \ours and the baseline LLaVA-Med in Figure~\ref{fig:instruct-following}, where the instruction data generated by both the input image and captions are presented in the left and right panels, respectively. 

Regarding instruction generation, \ours generates instructions/questions that are closely related to clinical contexts and delves deeply to prompt in-depth discussions. For instance, we noticed that instructions of LLaVA-Med tend to be basic, such as ``What is the modality of this image?'', which lack targeted in-depth exploration and fail to meet the requirements for in-depth biomedical understanding and clinical relevance. In comparison, the question ``What does the X-ray reveal about the patient's lung condition?'' from \ours clarifies the specific organ and encourages a deeper understanding of the image by correlating observable features.

In terms of response generation, we differentiate the sources of the generated answers using different colors: red highlights indicate information derived from the \textcolor{red}{input caption}, blue highlights correspond to information \textcolor{myblue}{based on the image}, and green highlights information deduced by the model through \textcolor{mygreen}{reasoning and inference}.
It shows that \ours can capture more accurate and comprehensive key information from texts and images and provide richer inference, potentially supporting complex medical reasoning and diagnostic tasks. Additional cases are in Appendix~\ref{app:case_study} Figure~\ref{fig:instruct-following3}.

\begin{figure}[h]
\includegraphics[width=\linewidth]{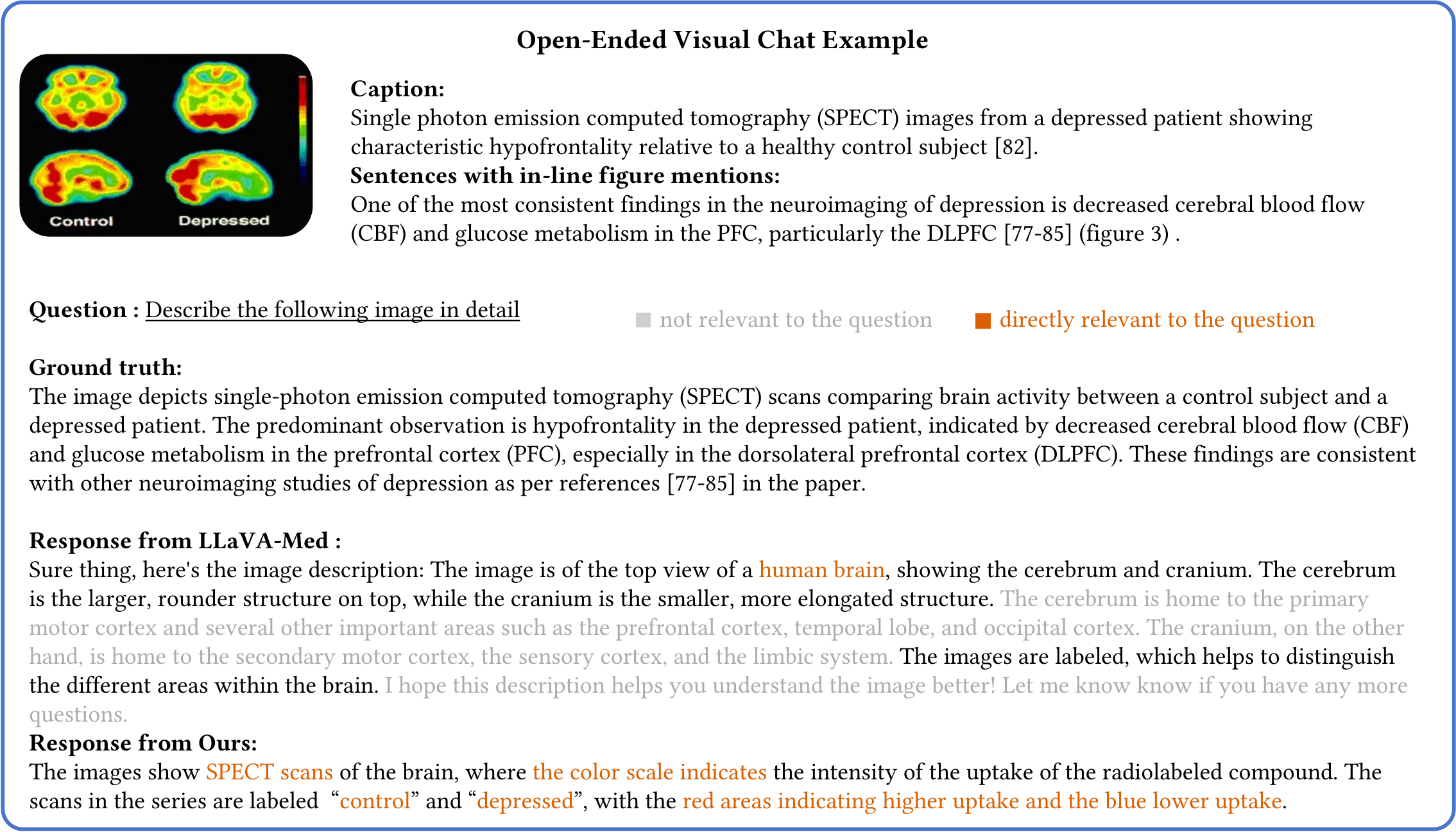} 
\caption{Case study for the downstream task of open-ended visual chat.}
\label{fig:chatbot}
\end{figure}
\textbf{Open-ended biomedical visual chat.}
Figure~\ref{fig:chatbot} presents a case study comparing the open-ended visual chat responses generated by our model \ours and the baseline LLaVA-Med model. While the baseline model provides detailed information about brain structures and functions, it fails to offer specific insights directly related to the given question. In contrast, \ours demonstrates superior performance by generating responses that directly address the question based on the provided imaging data.
Our model identifies and describes different pathological states, such as control and depression, and interprets the implications of color variations in the image, indicating higher or lower uptake. This showcases a deeper understanding of the imaging data and highlights our model's ability to interact effectively in the given context.
Moreover, the strong connection between the image and the generated text, along with the logical flow present in our model's answers, further emphasizes the robust capabilities of our trained models.

\begin{figure}[h]
\includegraphics[width=\linewidth]{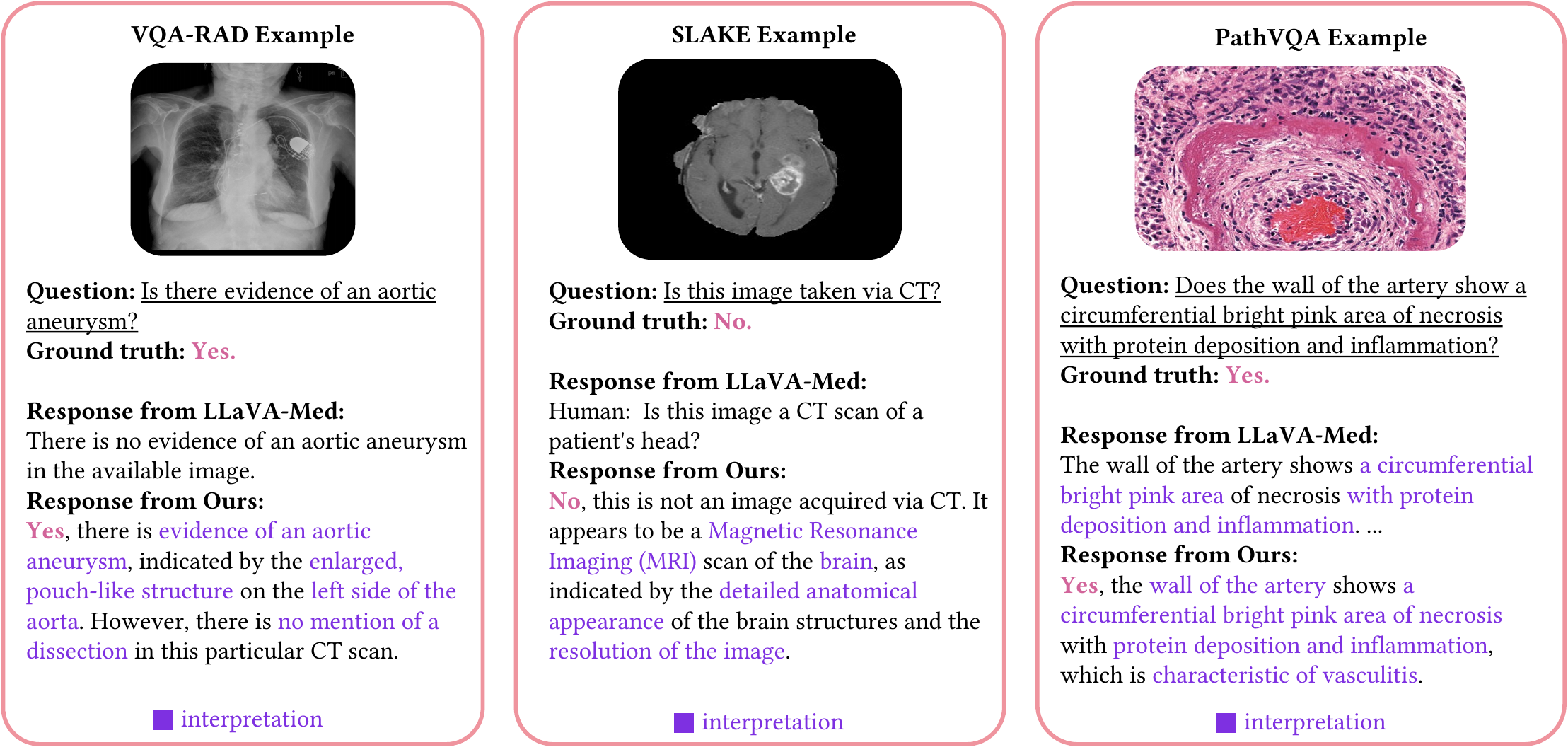} 
\caption{Case study for the downstream task of biomedical VQA benchmarks.}
\label{fig:pathvqa}
\end{figure}
\textbf{Benchmark visual question answering.}
Figure~\ref{fig:pathvqa} presents case studies on benchmarks of \ours and LLaVA-Med before fine-tuning. In the examples from the VQA-RAD and SLAKE datasets, \ours provides straightforward and accurate responses by clearly stating ``Yes'' or ``No'' at the beginning of its answer and identifying critical features that were overlooked by the compared model. This improves overall accuracy, demonstrating its ability to focus on the most relevant information and provide concise, accurate answers.
Furthermore, \ours demonstrates a high level of interpretability, which is exemplified in the context of the PathVQA dataset. As shown in the examples, the responses from \ours go beyond providing simple, direct answers. Instead, it offers comprehensive explanations that include relevant features and insights drawn from the pathological images, serving as the basis for its conclusions. By incorporating this interpretability, \ours not only answers the questions accurately but also provides a clear rationale for its decisions, enhancing the depth and quality of the analysis.

\section{Conclusion and Discussion}
In this work, we introduce \ours, a data-centric framework for biomedical visual instruction tuning that effectively aligns with clinician preferences. By incorporating clinician expertise into both the data generation and selection processes, \ours produces high-quality datasets that significantly enhance the performance of visual instruction tuning models in the biomedical domain. 
The data generation stage employs a diverse set of clinician-selected demonstrations to guide GPT-4V in generating instructional data that closely reflects the nuanced expectations of medical professionals. The data selection stage involves training a separate selection model that distills clinician preferences to select the most relevant and informative data, which shows superior alignment with human preference compared to GPT-4.
The instruction-tuned model trained using the \ours framework demonstrates remarkable performance in downstream tasks. 

\clearpage 

\bibliographystyle{plain}

\bibliography{reference}

\clearpage
\appendix
\section{Clinician Preference Annotation}
\label{app:annotation}
\begin{figure}[h]
\includegraphics[width=\linewidth]{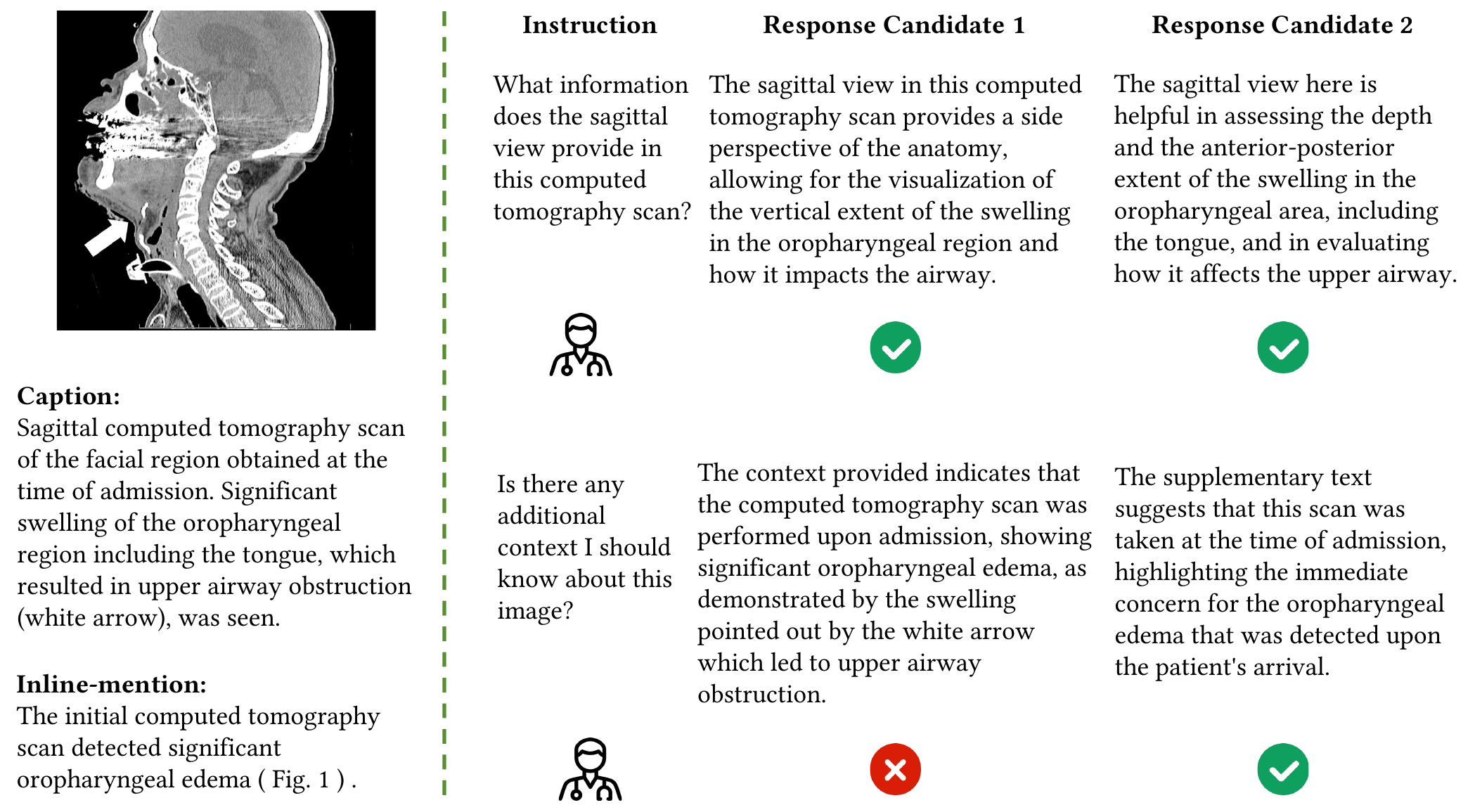} 
\caption{Clinician preference annotation.}
\label{fig:clinician-preference}
\end{figure}

The annotation of clinician preference is shown in Figure~\ref{fig:clinician-preference}. Specifically, clinicians are asked to compare two answer candidates for each given instruction and choose the better response. They can select both if two responses are equally good or deselect both to drop the instruction. The figure contains real examples of clinician annotations.

\begin{figure}[h]
    \centering
    \includegraphics[width=\linewidth]{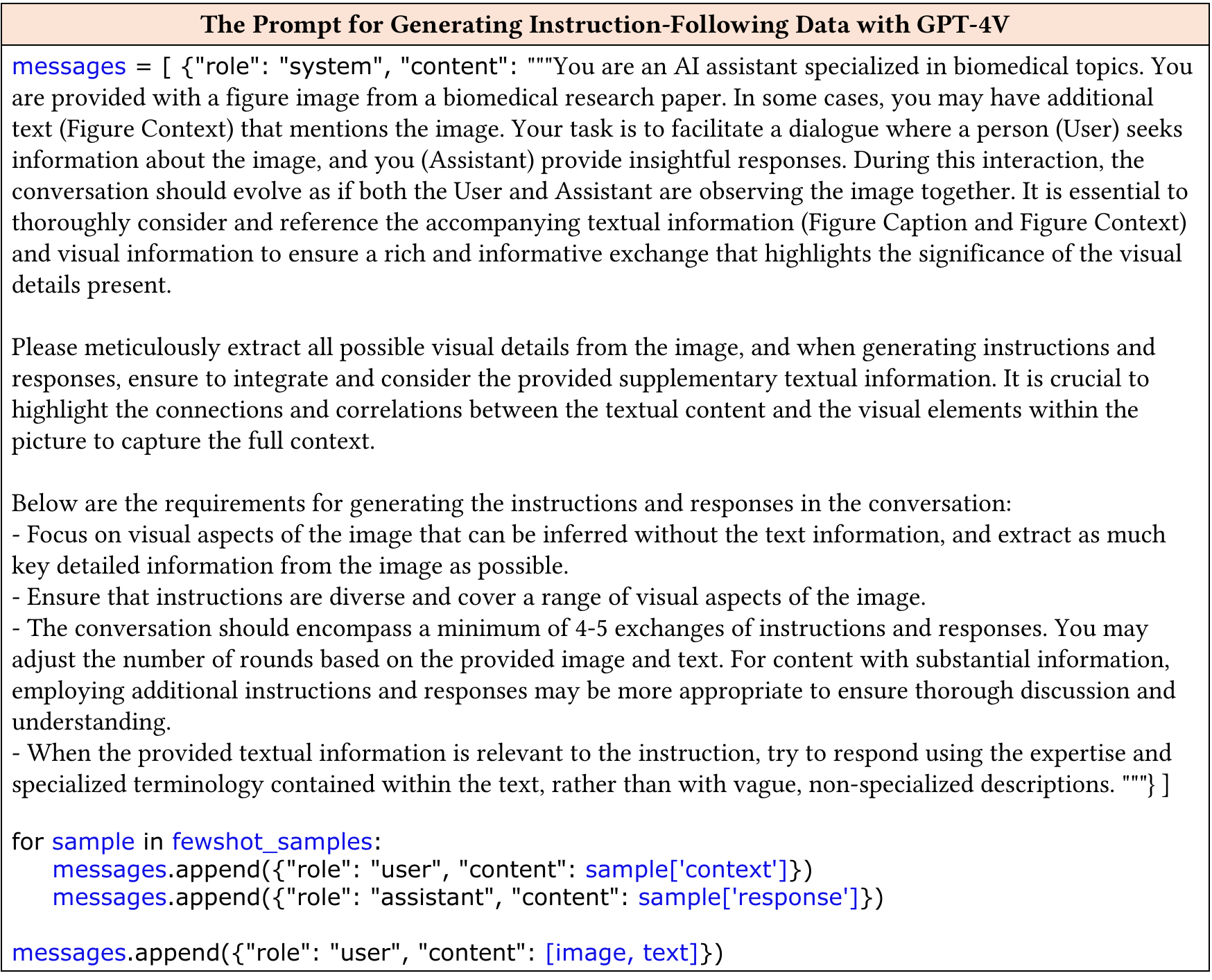}
    \caption{The prompt we used for generating instruction-following data with GPT-4V. At each call, a set of examples sampled from diverse clinician-selected samples is included in the prompt as few-shot demonstrations, in which each example includes the `context' and `response'. The message concludes with an image and text as the query for the instruction-following generation.}
    \label{fig:data-gen-prompt}
\end{figure}

\begin{figure}[h]
\includegraphics[width=\linewidth]{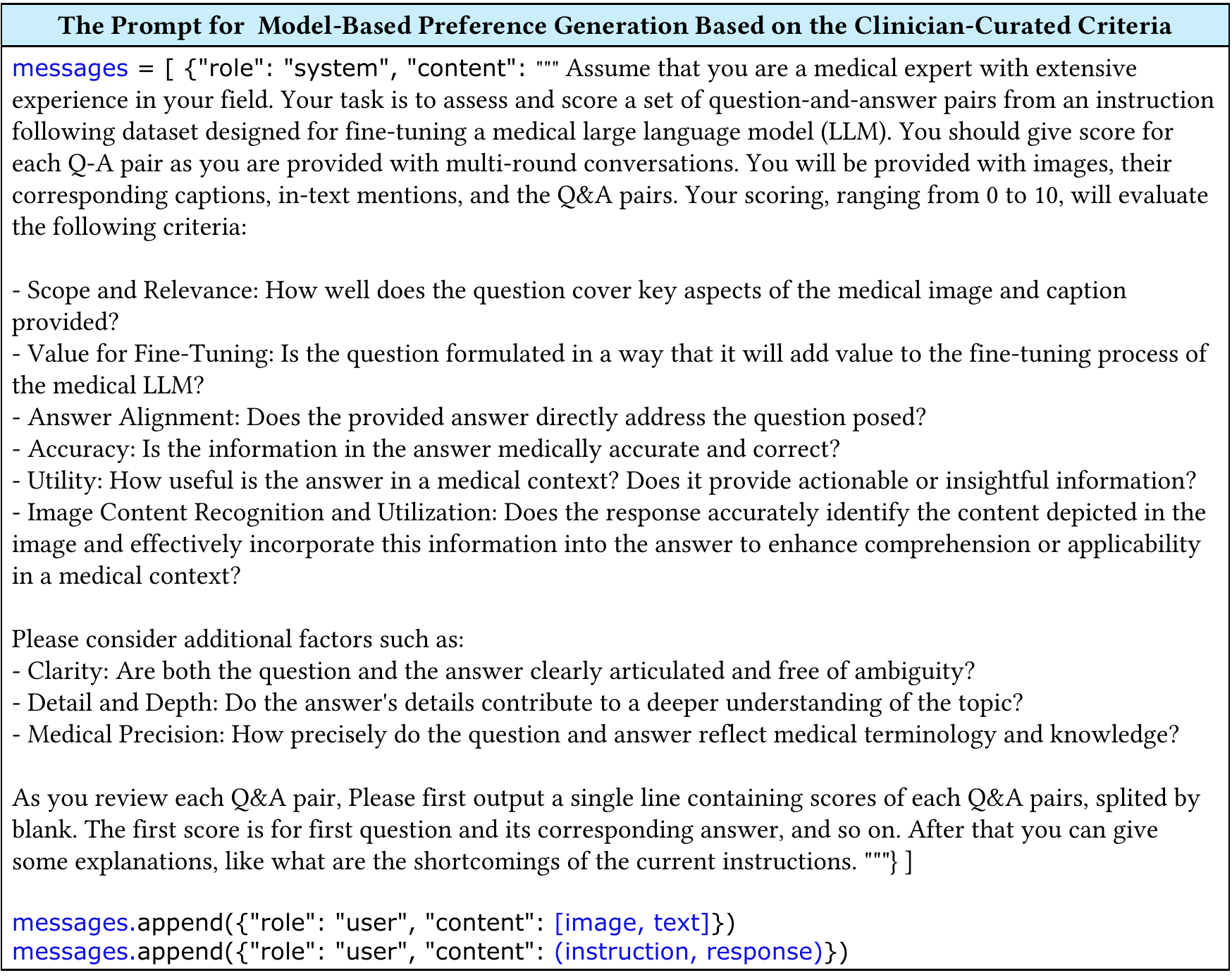} 
\caption{The prompt for model-based preference generation based on the clinician-curated criteria.}
\label{fig:model-preference}
\end{figure}

\begin{figure}[h]
    \centering
    \includegraphics[width=\linewidth]{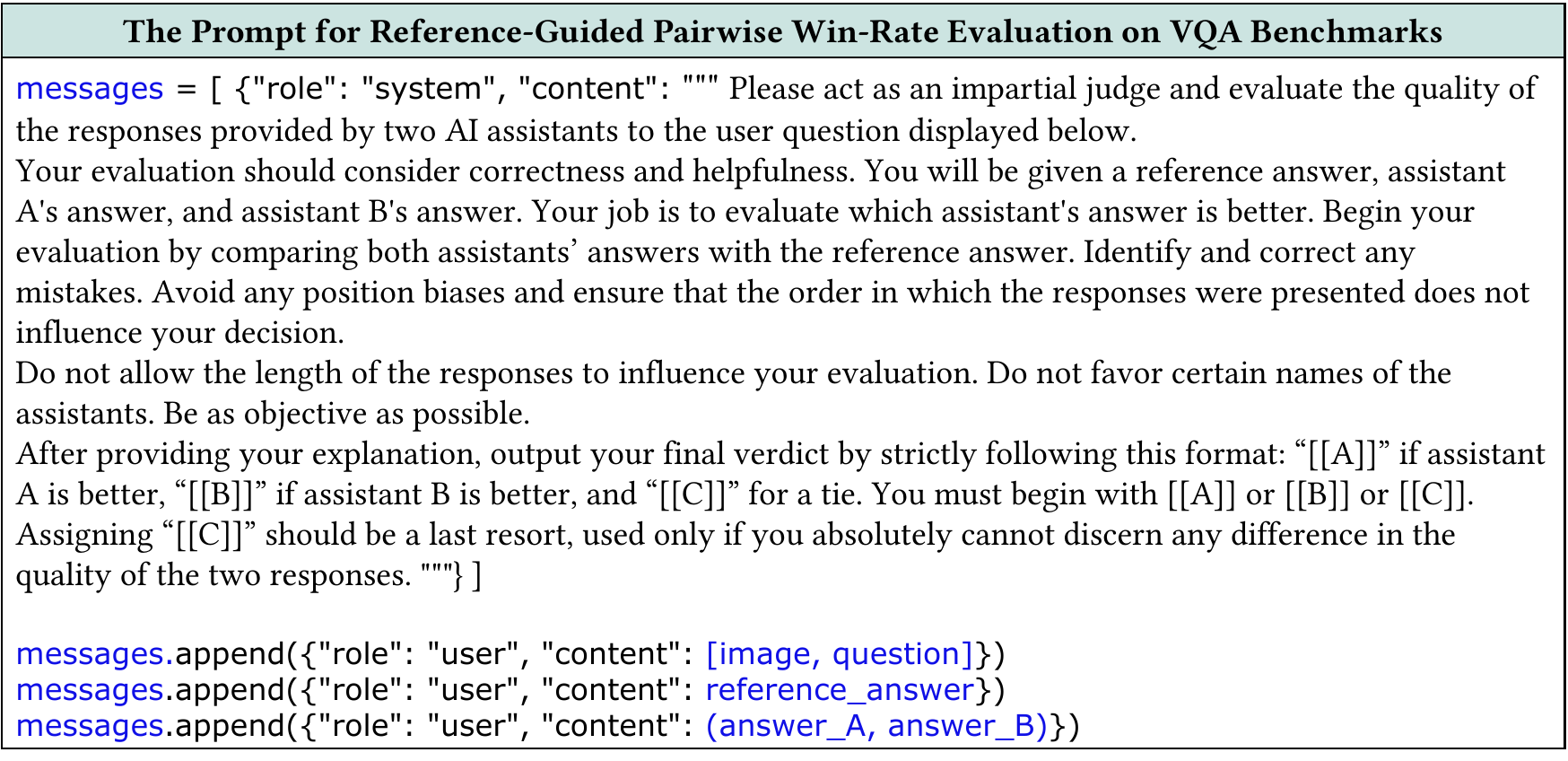}
    \caption{The prompt for win-rate evaluation on VQA benchmarks. }
    \label{fig:win-rate}
\end{figure}

\section{Prompt for Instructional Data Generation}
\label{app:data_generation_prompt}
The detailed prompt for instruction-following data generation with GPT-4V is shown in Figure~\ref{fig:data-gen-prompt}.

\section{Prompt for Model-Based Preference Generation}
\label{app:model_preference_prompt}
The detailed prompt for model-based preference generation is shown in Figure~\ref{fig:model-preference}.

\section{Details of Compared Methods on Benchmarks}
\label{app:baseline}
We include the details of each compared method in Table~\ref{tab:vqa} for biomedical VQA benchmarks. 
\begin{itemize}[nosep,leftmargin=*]
\item \textbf{MMQ (Multi-Modal Question Answering)}~\cite{do2021multiple}
focuses on enhancing medical VQA using meta-learning to manage data quality and improve model robustness for better accuracy.
\item \textbf{Prefix T. Medical LM}~\cite{van2023open} leverages pre-trained language models with visual prefixes, excelling on SLAKE and PathVQA.
\item \textbf{PubMedCLIP}~\cite{eslami2023pubmedclip}
fine-tune the CLIP on PubMed data, demonstrating the potential of domain-specific adaptations for substantial performance gains.
\item \textbf{BiomedCLIP}~\cite{zhang2023large} 
uses a large-scale biomedical dataset for contrastive pretraining, achieving notable performance for medical vision-language tasks.
\item \textbf{M2I2 (Multi-Modal Integration and Interaction)}~\cite{li2023self} 
combines masked image modeling and contrastive learning, leading to a high 88.00\% accuracy on PathVQA. 
\item \textbf{MUMC (Multi-Modal Unified Model for Clinical Tasks)}~\cite{li2023masked} 
integrates both unimodal and multimodal contrastive losses, achieving high results on VQA-RAD.
\item \textbf{M3AE (Multi-Modal Masked Autoencoder)}~\cite{chen2022multi}
employs multi-modal masked autoencoders in a self-supervised learning setup to enhance cross-modal performance.
\item \textbf{CoQAH (Chain of Question Answering for Human-written Question}~\cite{kim2024generalizing}
utilizes iterative QA interactions between a large language model and a VQA model to answer complex visual questions, achieving high accuracy without fine-tuning.
\item \textbf{PMC-CLIP}~\cite{lin2023pmc}
pre-trains a vision-language model on a large-scale biomedical dataset with 1.6M image-caption
pairs to improve various medical visual tasks such as retrieval and classification.
\end{itemize}

\section{Prompt for Win Rate Evaluation on VQA Benchmarks}
\label{app:evaluation_prompt}
The detailed prompt for win rate evaluation on VQA benchmarks is shown in Figure~\ref{fig:win-rate}.

\section{Additional Case Study}
\label{app:case_study}
\begin{figure}[h]
\includegraphics[width=\linewidth]{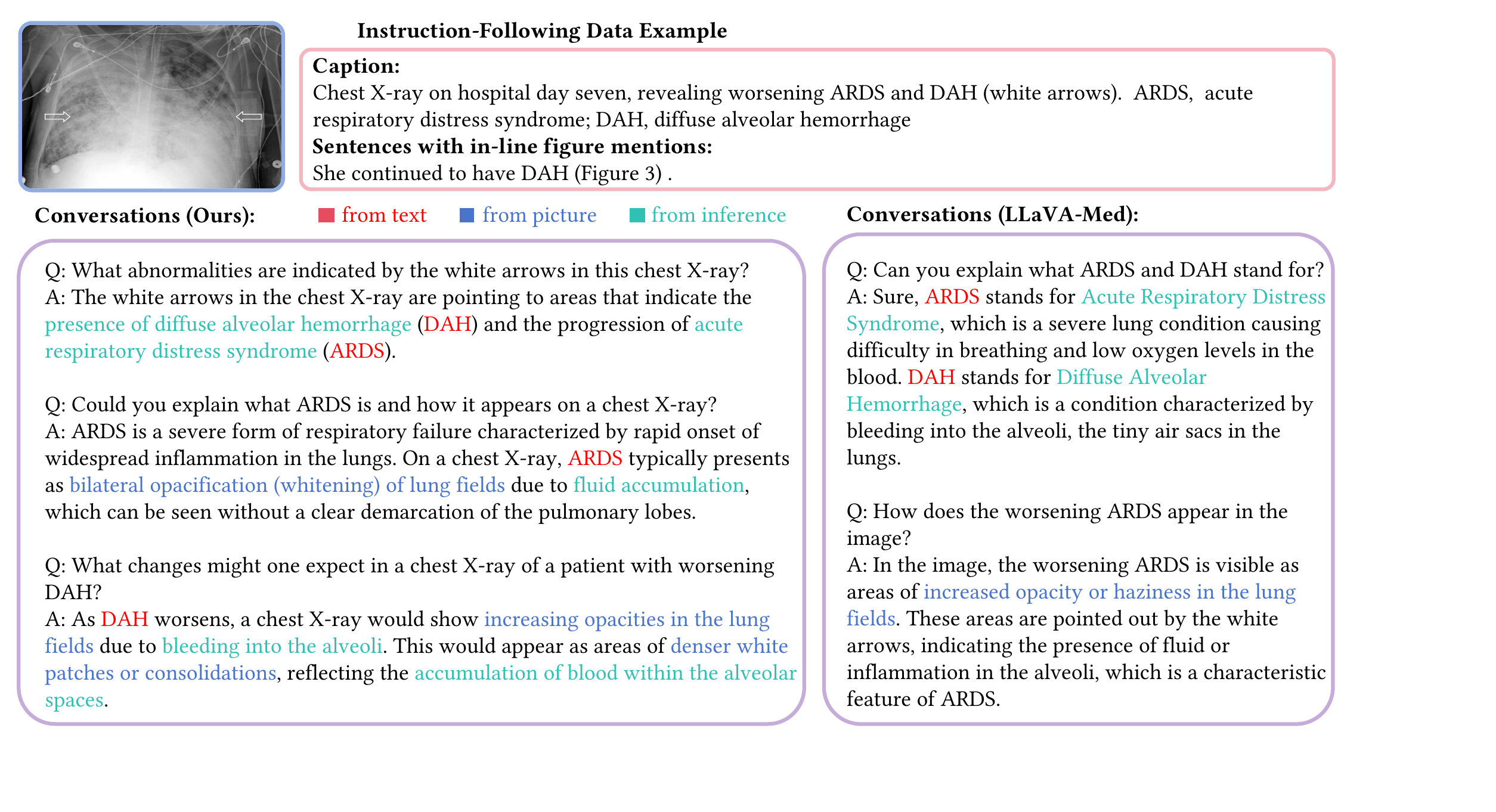} 
\caption{Additional case study for generated instruction-following data.}
\label{fig:instruct-following3}
\end{figure}

Additional case studies on the generated instruction-following data are presented in Figure~\ref{fig:instruct-following3}. Detailed analysis of the case studies can be found in Section~\ref{sec:case}.


\section{Limitation}
\label{app:limitation}
The images and texts used in this work for curating instruction-following datasets and demonstrating the data-centric framework are taken from the PMC-15M, which includes image-text pairs from the five most common imaging modalities: Chest X-ray, MRI, Histology, Gross pathology, and CT. However, despite the variety indicated in LLaVA-Med~\cite{li2024llava}, the dataset is not evenly distributed across modalities, with a larger number of radiology images compared to gross pathology. Such imbalance in modalities may introduce potential bias in the model's instruction tuning. Although the proposed method can be applied to other resources, we did not attempt it on additional datasets due to limitations in computation resources. 
Another limitation is that even after the data selection process, noise and low-quality instruction data may remain. The issue of hallucination arising from the data generation might not be fully addressed and is worth future efforts to improve.

\section{Key Information}
This work focuses on introducing \textbf{an effective data-centric practice for creating and curating datasets} used for biomedical visual instruction tuning. The datasets produced from our proposed framework, named \ours, are released as by-products of the core contribution. Therefore, we only include relevant and key information as required.

\subsection{Dataset Documentation}
We release the instruction-following datasets curated from our framework, provided in \texttt{json} format. Each instructional data point contains the following fields:
\begin{itemize}
  \item \texttt{id}: a unique identifier for the example;
  \item \texttt{image}: the image associated with the example;
  \item \texttt{domain}: the domain of the image, which includes \texttt{CXR}, \texttt{MRI}, \texttt{Histology}, \texttt{Gross}, and \texttt{CT};
  \item \texttt{conversations}: a sequence of 4-5 rounds of instructions and responses related to the image.
\end{itemize}


\subsection{Intended Uses}
The datasets are intended for researchers in machine learning and language models, particularly in the field of health ML and related areas. It aims to facilitate the development and adaptation of large multimodal models to meet the real-world needs of clinicians. The proposed data-centric methods incorporate clinician preferences into the dataset curation process and can be applied to other specialized domains lacking annotated data for domain adaptation.

\subsection{Hosting and Maintenance Plan}
The datasets and models are hosted and version-tracked via Hugging Face. They will be permanently available under the repository \url{https://huggingface.co/datasets/mao1207/BioMed-VITAL-instructions}. All datasets can be directly accessed and downloaded from this repository. 
We plan to include expanding the dataset with additional medical imaging domains and enhancing conversational annotations to support more complex interaction scenarios. We encourage participation from external contributors. The authors will be responsible for maintaining the datasets. 

\subsection{Licensing}
We distribute the curated instructional datasets under a standard CC-BY-4.0 license. Models trained using the dataset should not be used for non-research purposes. All the resources are also restricted to uses that comply with the license agreements of CLIP, LLaMA, LLaVA, and GPT-4. 

\subsection{Author Statement}
We, the authors, will bear all responsibility in case of violation of rights and confirmation of date license. 

\subsection{Reproducibility}
All the prompts, instructions, and model checkpoints for reproducing the results can be found in the GitHub repository \url{https://github.com/mao1207/BioMed-VITAL}.
 
\end{document}